\documentclass{article}
\usepackage{ijcai21}

\usepackage{times}
\usepackage{soul}
\usepackage{url}
\usepackage{enumitem}
\usepackage[hidelinks]{hyperref}
\usepackage[utf8]{inputenc}
\usepackage[small]{caption}
\usepackage{graphicx}
\usepackage{amsmath}
\usepackage{amsthm}
\usepackage{booktabs}
\usepackage{algorithm}
\usepackage{algorithmic}
\usepackage{multirow}
\usepackage{subcaption}
\title{AdaVQA: Overcoming Language Priors with Adapted Margin Cosine Loss\thanks{Corresponding Author: Liqiang Nie and Zhiyong Cheng.}}

\author{
Yangyang Guo$^1$
\and
Liqiang Nie$^1$
\and
Zhiyong Cheng$^2$
\and
Feng Ji$^3$
\and
Ji Zhang$^3$
\and
Alberto Del Bimbo$^4$
\affiliations
$^1$Shandong University
$^2$Shandong Artificial Intelligence Institute
$^3$Alibaba Group
$^4$University of Florence
\emails
\{guoyang.eric,  nieliqiang,  jason.zy.cheng\}@gmail.com,
\{zhongxiu.jf, zj122146\}@alibaba-inc.com,
alberto.delbimbo@unifi.it
}

\begin{document}

\maketitle

\begin{abstract}
A number of studies point out that current Visual Question Answering (VQA) models are severely affected by the language prior problem, which refers to blindly making predictions based on the language shortcut. Some efforts have been devoted to overcoming this issue with delicate models. However, there is no research to address it from the angle of the answer feature space learning, despite of the fact that existing VQA methods all cast VQA as a classification task. Inspired by this, in this work, we attempt to tackle the language prior problem from the viewpoint of the feature space learning. To this end, an adapted margin cosine loss is designed to discriminate the frequent and the sparse answer feature space under each question type properly. As a result, the limited patterns within the language modality are largely reduced, thereby less language priors would be introduced by our method. We apply this loss function to several baseline models and evaluate its effectiveness on two VQA-CP benchmarks. Experimental results demonstrate that our adapted margin cosine loss can greatly enhance the baseline models with an absolute performance gain of 15\% on average, strongly verifying the potential of tackling the language prior problem in VQA from the angle of the answer feature space learning.
\end{abstract}

\section{Introduction}\label{introduction}
\begin{figure*}
  \centering
  \begin{minipage}{0.45\linewidth}
  \centering
  \includegraphics[width=1.0\linewidth]{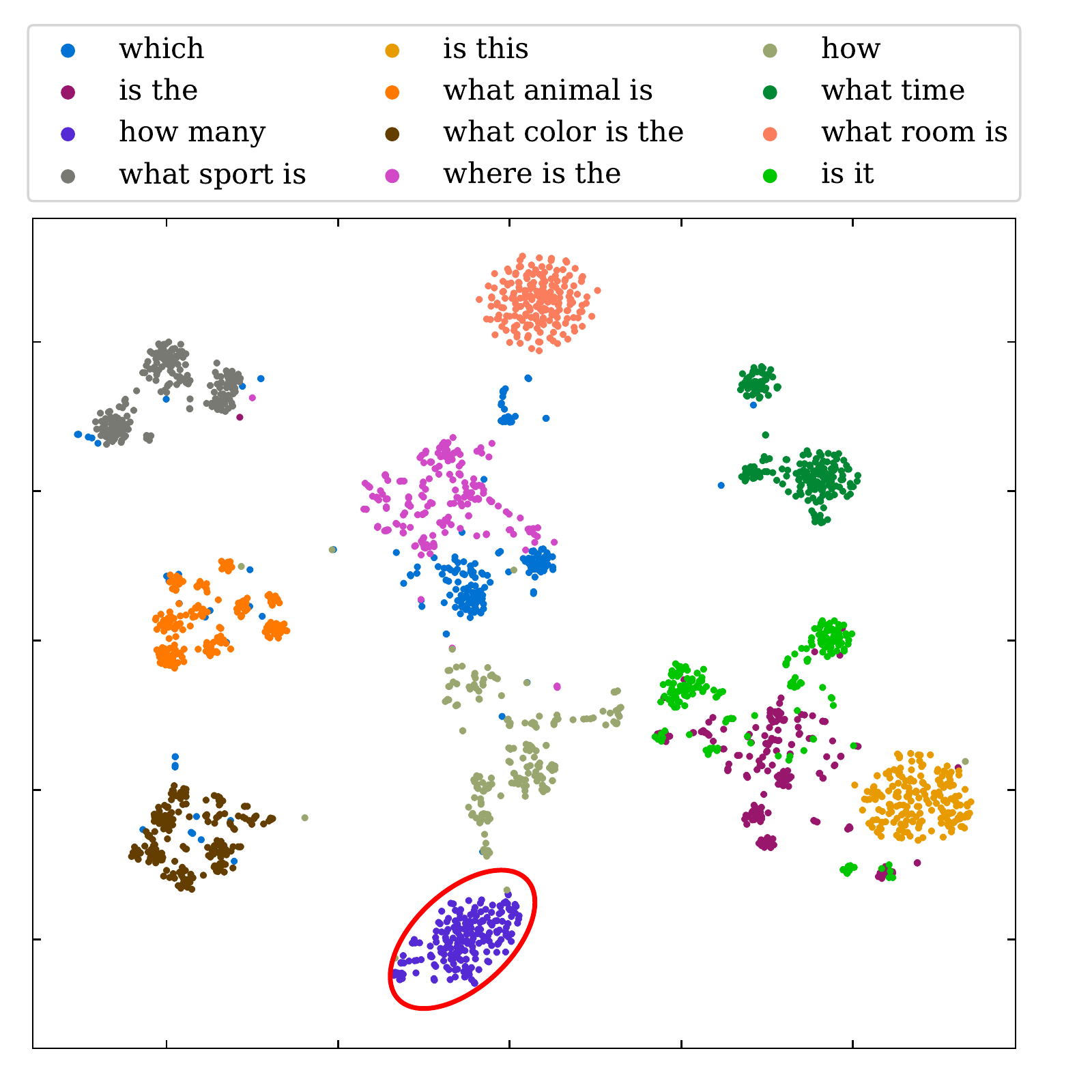}
  \subcaption{Answer manifold embedding with respect to question types.}\label{fig:embed-type}
  \end{minipage}
  \begin{minipage}{0.53\linewidth}
  \centering
  \begin{minipage}{0.9\linewidth}
  \centering
  \includegraphics[width=1.0\linewidth]{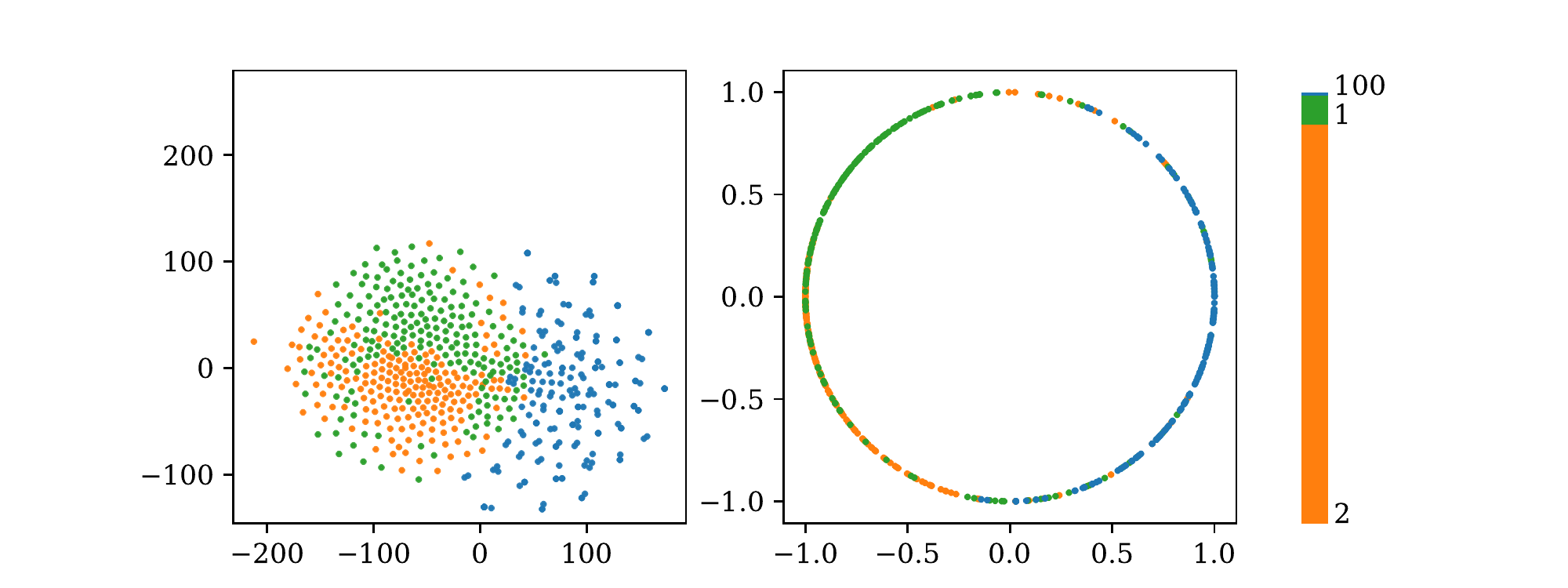}
  \subcaption{Answer manifold embedding of \emph{how many} question type in Euclidean and angular spaces from the baseline. }\label{fig:embed-base}
  \end{minipage}

  \begin{minipage}{0.9\linewidth}
  \centering
  \includegraphics[width=1.0\linewidth]{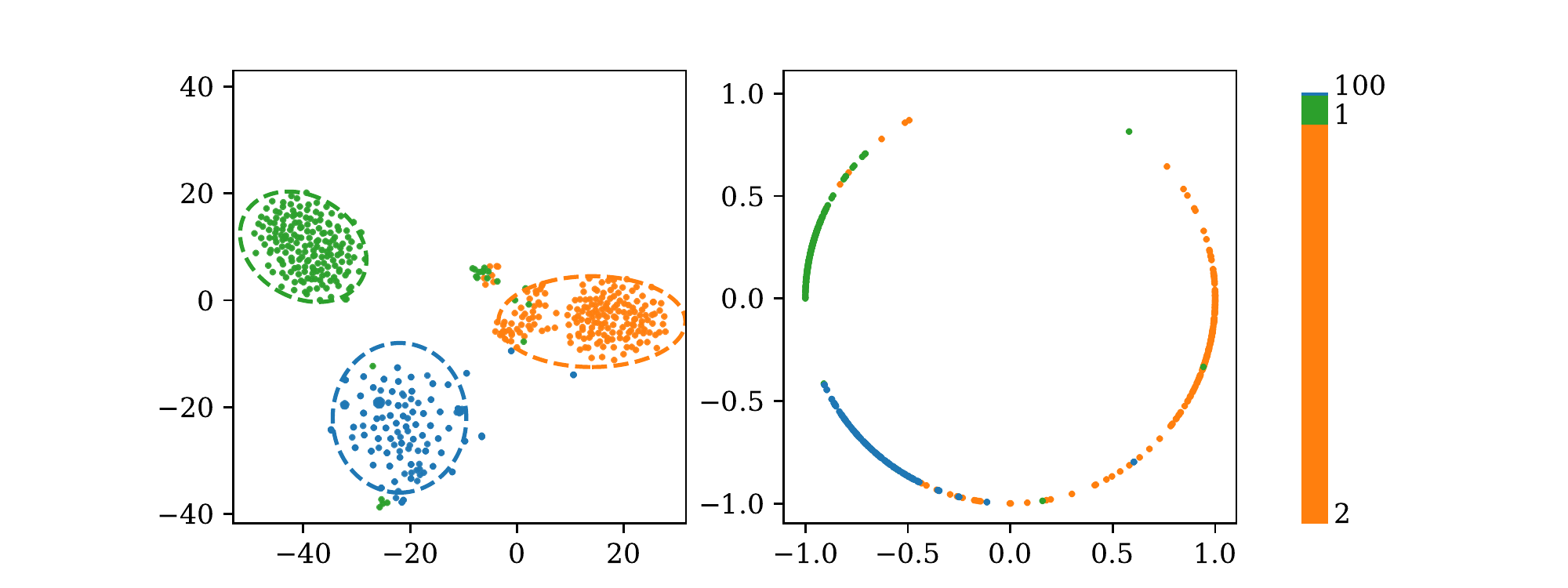}
  \subcaption{Answer manifold embedding of \emph{how many} question type in Euclidean and angular spaces from our method. } \label{fig:embed-our}
  \end{minipage}
  \end{minipage}
  \caption{Answer manifold embedding with respect to questions types in (a) and answers under \emph{how many} question type in (b) and (c). Note that the numbers of each answer class are identical for better visualization. For (b) and (c), the answer distribution in the train set is demonstrated on the rightmost.}\label{fig:embed}
\end{figure*}

Visual Question Answering (VQA) targets at accurately answering natural language questions regarding a visual scene. Perceived as a more `AI-complete' task, VQA has received critical attention from both computer vision and natural language processing communities. Recently, several large-scale benchmarks~\cite{vqa1,vqa2} have been constructed to facilitate the advancement of VQA, followed by a number of elaborately designed visual reasoning
models~\cite{strong,updown,count}.

Despite its flourishing development, there still exits a major issue in VQA, i.e., most VQA models tend to make predictions by taking the language shortcut between questions (or more specifically, question types\footnote{We refer question type in VQA to the first few words in the given question.}) and answers. For example, \emph{2} takes large proportions of the answers to questions initiated with \emph{how many} in the train set. And then the model leverages these language priors and directly yields answer \emph{2} to \emph{how many} questions when testing, regardless of both the remaining question words and the given image. As a result, the visual signal is largely under-explored, making the visual reasoning in VQA deteriorate into the a pure language matching problem without the consideration of `V'. Obviously, it is a contradiction to the initial motivation of this vision-and-language task.

Many research efforts have been devoted to overcoming this arduous problem, which can be roughly classified into two groups: 1) balancing the biased dataset; and 2) correcting VQA models for alleviating the effects of language priors. In detail, the first category of methods contribute to either enlarging~\cite{vqa2} or re-splitting~\cite{vqacp} the existing biased datasets. Notably, ~\cite{vqacp} curated a diagnostic dataset - VQA-CP, wherein the answer distributions of per question type are significantly distinct between the train and test sets. It thus provides a favorable test-bed for estimating the language prior effect the VQA model encounters. Based on this dataset, some pioneering work has been conducted to correct the advanced VQA models for alleviating language prior problem~\cite{adversarial-nips,counterfactual,loss-vqa}. A typical design of this category of methods is to attach an additional question-only branch, whereby the language prior is intentionally captured and further suppressed by the orthodox question-image branch.

Due to the fact that the answers in most VQA datasets are composed of a few words, existing VQA models often cast VQA as a classification task, i.e., classifying the answer class rather than progressively generating answer words to achieve better visual reasoning capability. However, what attracts our attention is that, there has been no research diving into investigating the learned feature space that each answer/class engages. We argue that the feature space learned from the biased dataset cannot distinguish the answers distinctly and is the key to alleviate the language prior effect. Inspired by this thought, we empirically visualize the 2-D embedding features of the answers to the question type \emph{how many} in Fig~\ref{fig:embed-base}. One can see that different answers of the baseline are actually intertwined with each other in the two kinds of feature spaces. On the basis of this phenomenon, we make a further assumption: is it beneficial for overcoming the language prior problem if we properly separate the answers via manipulating their learned feature space?

To address this question, we propose to revisit the language prior problem from the viewpoint of the answer feature space learning. The primary goal of this work is to introduce an adapted margin for different answers in the angular space under the corresponding question type, which can effectively separate the answer embeddings. Towards this end, we firstly reformulate the softmax loss function as a cosine loss by $L2$ normalizing both answer features $\mathbf{x}$ and weight vectors $\mathbf{W}_i$~\cite{asoftmax,cosface}. Consequently, the decision boundary can be computed through the cosine function of the angle $\theta_i$ between $\mathbf{x}$ and $\mathbf{W}_i$, i.e., $\mathbf{W}_i^T \mathbf{x} = ||\mathbf{W}_i||_2 \cdot ||\mathbf{x}||_2 \cdot cos \theta_i = cos\theta_i$, s.t. $||\mathbf{W}_i||_2=1$ and $||\mathbf{x}||_2 = 1$. Thereafter, an adapted margin $m_i$ is employed to separate the answer features, i.e., $cos\theta_i - m_i$, where $m_i$ is computed based on the train set statistics of answer $a_i$ under the corresponding question type. In the ideal scenario, for each given question and its corresponding question type, frequent answers would span broader in the angular space with a smaller margin, while sparse answers span tighter in the angular space with a larger margin (see Fig.~\ref{fig:embed-our}).  The intuition behind this is that frequent answers under certain question types often go with more training samples, and thus a smaller margin is required to control a broader space to sufficiently include these answers. In contrast, the number of sparse answers is relatively small, requiring a more tighter feature space. Moreover, we find that the answer embeddings in the Euclidean space become more discriminative, and further benefit the reducing of the language prior effect.

In a nutshell, the proposed adapted margin cosine loss is beneficial in overcoming the language prior problem in VQA, making itself being model-agnostic and easy to implement. Besides, it introduces NO incremental parameters or computational cost when testing. To the best of our knowledge, this is the first attempt to directly optimize the feature space learning to attack this problem. Additionally, we apply this loss function over three baseline methods, verifying its effectiveness on two VQA-CP benchmarks. The experimental results demonstrate that the baselines equipped with our loss achieve an absolute gain of over \textbf{15}\% averagely. The code is released for the re-implementation of this work\footnote{https://github.com/guoyang9/AdaVQA.}.

\section{Related Work}\label{related_work}
In recent years, a considerable literature has thrived around the task of VQA~\cite{han,vqa1,vqa2,fvqa,count}. From the very beginning, researchers proposed to encode the question and image into a common latent space, based on which the accurate answers can be identified through classification~\cite{vqa1,ask}. With the prevailing advent of the attention mechanism, different contributions of each image region~\cite{san,mcb,strong,updown} or each question word~\cite{hierarchical,mfb} for the answer prediction have been taken into consideration. Some studies have also explored the modular structure of questions for more explicit reasoning~\cite{nmn}. Nevertheless, questions in real world often require reasoning beyond what is in the image. To better handle such situation, FVQA~\cite{fvqa} and OK-VQA~\cite{okvqa} develop new benchmarks as well as insightful approaches to incorporate structured and unstructured knowledge in VQA, respectively.

It is noted that many VQA models are influenced by the language prior problem~\cite{vqacp,lpscore}. Therefore, some efforts have been devoted to the biased dataset balancing. For instance, VQA v2~\cite{vqa2} adds complementary samples to the original VQA v1 dataset~\cite{vqa1} such that for each sample $<$\emph{v,q,a}$>$ , another one with the same \emph{q}, similar \emph{v} yet a different \emph{a} is generated. VQA-CP~\cite{vqacp} re-splits the VQA v1 and v2 datasets that the answer distribution of per question type is distinct between the train and test sets. In addition to the studies on the dataset side, methods devised for tackling this problem mostly try to diminish the answer prediction influence from the question encoding layer~\cite{hint,critical,vgqe,questiontype}. One popular solution essentially involves a new question-only training branch, wherein the answers are predicted based solely on the question input~\cite{adversarial-nips,rubi,lmh,counterfactual}. The answer prediction from this branch is further suppressed by the original question-image one. 
\section{Proposed Method}\label{model}
\subsection{Problem Definition}
Following the prevalent formulation, we consider the VQA task as a multi-label classification problem. That is, for a textual question $Q$ regarding an image $I$, the objective function is given by:
\begin{equation} \label{equ:definition}
    \hat{a} = \mathop{\arg \max}_{a \in \Omega}  p(a|Q, I; \Theta),
\end{equation}
where $\Omega$ and $\Theta$ denote the answer set and the model parameters, respectively. Note that there can be multiple correct answers for each instance. The negative log likelihood loss function is then formulated as\footnote{The individual loss is considered instead of the total instances throughout this paper due to the space limitation.}:
\begin{equation} \label{equ:softmax}
\begin{aligned}
    L_s &= \sum_{i=1}^{|\Omega|} - a_i \log p_i \\
        &= \sum_{i=1}^{|\Omega|} - a_i \log \frac{\exp (\mathbf{W}_i^T \mathbf{x})}{\sum_{j=1}^{|\Omega|} \exp (\mathbf{W}_j^T \mathbf{x})},
\end{aligned}
\end{equation}
where $\mathbf{W}$ and $\mathbf{x}$ denote the weight matrix and feature vector directly adjacent to the answer prediction, respectively; $a_i \in [0, 1]$ represents the corresponding answer label. Besides, we remove the bias vector for simplicity as we also found it contributes little to the final model performance.

\subsection{Method Intuition}
To date, most VQA methods tackling the language prior problem present to confront the prediction from a question-only branch. Nevertheless, there is no literature exploring the answer/class embedding in the learned feature space. Therefore, in this work, we adopt a new viewpoint to overcome this problem, and validate its effectiveness in what follows. Specifically, we propose to place an adapted margin to separate the answer embeddings in the angular space properly based upon the train set statistics. Compared to the Euclidean space, the angular space is less complicated to manipulate, as the margin range reduces from $(-\infty, +\infty)$ to $[-1, 1]$. In the light of this, we firstly leverage the $L2$ normalization on weight vector $\mathbf{W}_i$ and feature vector $\mathbf{x}$ to ensure the posterior probability to be determined by the angle $\theta_i$ between $\mathbf{W}_i$ and $\mathbf{x}$ (the answer feature space is converted from the Euclidean space to the angular space), i.e., $||\mathbf{W}_i||_2=1$ and $||\mathbf{x}||_2=1$. Accordingly, a modified normalized SoftMax loss (NSL)~\cite{cosface} can be obtained by,
\begin{equation} \label{equ:ns}
    L_{ns} = \sum_{i=1}^{|\Omega|} - a_i \log \frac{\exp{s \cos{\theta_{i}}}}{\sum_{j=1}^{|\Omega|} \exp{s \cos{\theta_j}}},
\end{equation}
where $s$ is a scale factor for more stable computation. In order to achieve a more discriminative classification boundary, LMLC~\cite{cosface} introduces a fixed cosine margin to NSL,
\begin{equation} \label{equ:nsl}
\resizebox{1.0\hsize}{!}{$
    L_{lmc} = \sum_{i=1}^{|\Omega|} - a_i \log \frac{\exp{s ( \cos{\theta_{i}} - m)}}{\sum_{j \neq i} \exp{s\cos{\theta_j}} + \exp{s (\cos{\theta_{i}} - m)}},
    $}
\end{equation}
where $m$ implies the fixed cosine margin.

However, applying a fixed cosine margin cannot obtain satisfactory results as observed in our experiments. For this situation, the key reason is that the answer distribution of each question type is highly biased, resulting in the incapability of learning a sufficient representation with a fixed margin in the angular space. In the next subsection, we will introduce a more sophisticated adapted margin cosine loss to overcome this issue.
\begin{figure}
  \centering
  \includegraphics[width=0.9\linewidth]{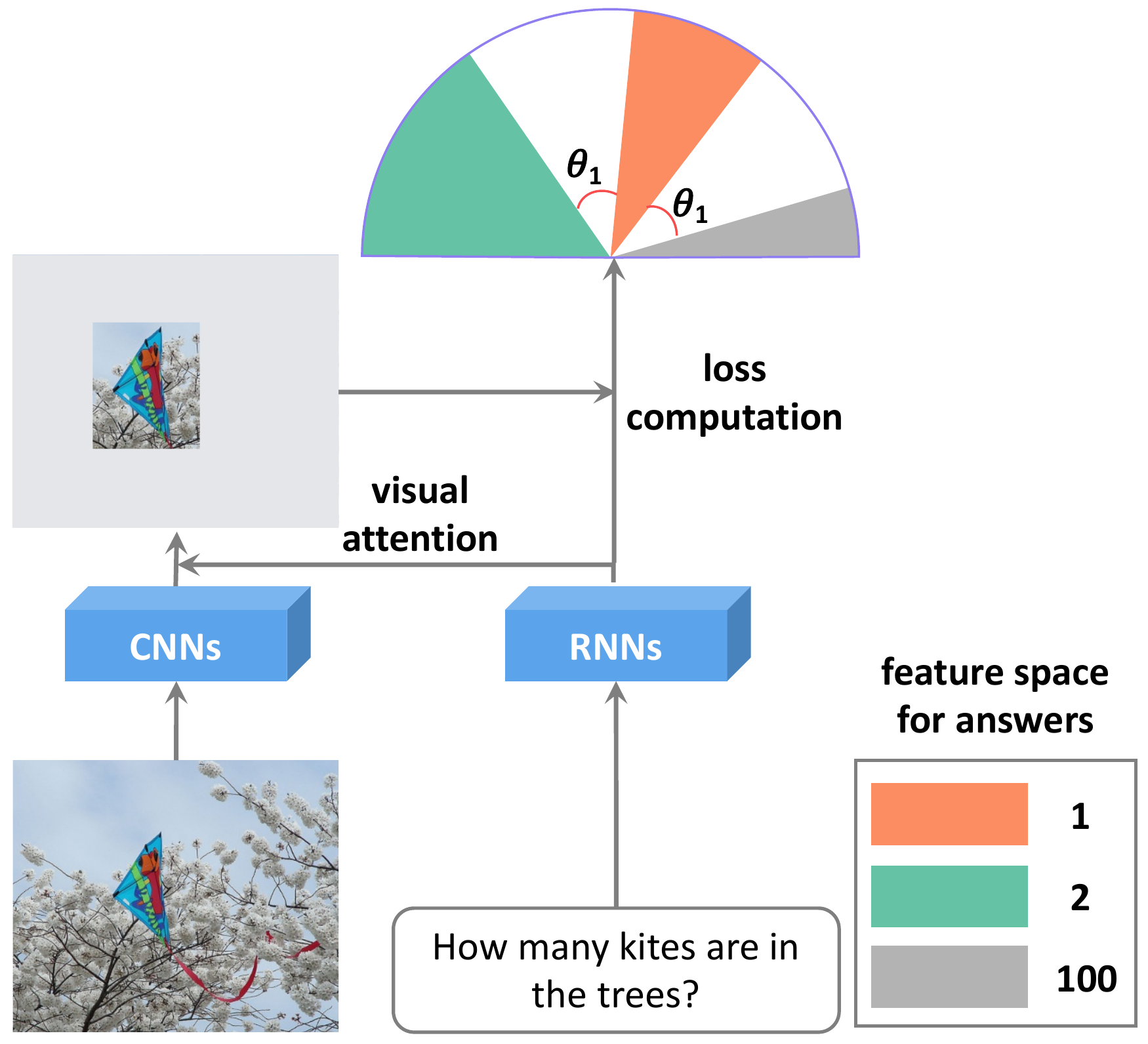}
  \caption{The training pipeline of AdaVQA. The input image and question is consumed via pre-trained convolutional neural networks (CNNs) and recurrent neural networks (RNNs), respectively. A visual attention module is then postfixed for identifying salient image regions pertaining to the given question. After normalizing the weight and feature vectors, an adapted cosine margin is introduced to produce suitable decision boundaries.}\label{fig:pipeline}
\end{figure}

\subsection{AdaVQA Formulation}
The results in our experiments (see Sect.~\ref{exp:ablation}) explicitly demonstrate that a fixed cosine margin yields limited improvements or even deteriorates the model performance. Based upon this observation, we argue that an adapted cosine margin is more favorable for overcoming the language priors in VQA. In view of this, a new loss function is elaborately defined as:
\begin{equation}
\left\{
\begin{aligned}
    & L_{AdaVQA} = \sum_{i=1}^{|\Omega|} - a_i \log \frac{\exp{s (\cos{\theta_{i}} - m_i)}}{\sum_{j=1}^{||\Omega||} \exp{s (\cos{\theta_j} - m_j)}},
      \\
    & m_i = 1 - \bar{m}_i, \\
    & \bar{m}_i = \frac{n_i^k + \epsilon}{\sum_{j=1}^{|\Omega|}n_j^k + \epsilon},
\end{aligned}
\right.
\end{equation}
where $m_i$ is the adapted margin for answer $i$ based on the given question type, $n_i^k$ denotes the number of answer $i$ under the question type $qt_k$ in the train set, $\epsilon = 1e-6$ is a hyper-parameter for avoiding computational overflow. The intuition behind this is that, for the current given question and its corresponding question type, the frequent answers span broader in the angular space (smaller margin), while sparse answers span tighter (larger margin). In other words, frequent answers imply more training samples, which require a broader feature space to sufficiently include these answers. On the contrast, a tighter feature space is acceptable for sparse answers as the number of training samples is much smaller. This setting enables the VQA models to place a more proper cosine margin in the angular answer feature space. Taking Fig.~\ref{fig:embed-our} as an example, answer \emph{2} is more frequent than the other two answers (\emph{1} and \emph{100}) in the train set, which results in a larger feature space as we expect. In particular, Fig.~\ref{fig:pipeline} provides a general training pipeline of AdaVQA, where the feature space for the three answers (\emph{1, 2, 100}) is learned based on the proposed loss function.

However, one may concern that such a margin restriction might damage the inter-type discrimination, i.e., the answer decision boundary among different question types becomes less discriminative. We therefore obtain the answer embeddings with respect to its question type and visualize them in Fig.~\ref{fig:embed-type}. From this figure, an obvious decision boundary pertaining to question types can be observed. One possible reason to this situation is that the cosine margin of each answer is computed according to the train set statistic under the given question type, while other answers outside this question type are kept untouched. This empirical setting inherently equip the model with the discrimination capability among different question types.

\textbf{Entropy Threshold.}
One major problem have yet to be addressed, i.e., not all questions within their corresponding question type would cause the language prior problem. Different from previous methods handling all questions without differentiation, we propose to employ an entropy threshold over each question type. In particular, the questions whose question type entropy is greater than the threshold should be considered for regularization. Specifically, the entropy for each question type is defined as:
\begin{equation}
    e_{qt} = - \sum_{i=1}^{|\Omega|} \bar{m}_i^k \log_2 \bar{m}_i^k,
\end{equation}
where $\bar{m}_i^k$ is actually the occurrence probability of answer $i$ under the question type $qt_k$ in the train set.

\textbf{Partial Derivatives.} We further provide the partial derivatives of the weight vector $\mathbf{W}_i$ and feature vector $\mathbf{x}$ from our loss function. Let $p_i = s (\cos{\theta_{i}} - m_i) = s(\frac{\mathbf{W}_i^T}{||\mathbf{W}_i||_2} \cdot \frac{\mathbf{x}}{||\mathbf{x}||_2} - m_i)$, and $\hat{p}_i = \frac{\exp{p_i}}{\sum_{j=1}^{|\Omega|}\exp{p_j}}$. The partial derivatives are obtained via,
\begin{equation}
\resizebox{1.0\hsize}{!}{$
    \frac{\partial L_{AdaVQA}}{\partial \mathbf{x}} =  \sum_{i=1}^{|\Omega|} (\sum_{j=1}^{|\Omega|} a_j \hat{p}_i - a_i) \cdot s \cdot  \frac{||\mathbf{x}||_2 \mathbf{I} - \mathbf{x}\mathbf{x}^T}{||\mathbf{x}||_2^3} \frac{\mathbf{W}_i}{||\mathbf{W}_i||_2},
$}
\end{equation}
and
\begin{equation}
\resizebox{1.0\hsize}{!}{$
    \frac{\partial L_{AdaVQA}}{\partial \mathbf{W}_i} =  \sum_{i=1}^{|\Omega|} (\sum_{j=1}^{|\Omega|} a_j \hat{p}_i - a_i) \cdot s \cdot  \frac{||\mathbf{W}_i||_2 \mathbf{I} - \mathbf{W}_i \mathbf{W}_i^T}{||\mathbf{W}_i||_2^3} \frac{\mathbf{x}}{||\mathbf{x}||_2}.
$}
\end{equation}

\textbf{Lower Bound for s.} The scale factor $s$ is critical for the final answer feature learning. A too small $s$ leads to an insufficient convergence as it limits the feature space span (as we found in our experiments that the loss goes `nan' with a small $s$) In view of this, a lower bound for $s$  should be prescribed. Without loss of generality, let $P_{i}$ denote the expected minimum of the answer class $i$. Therefore, we have,
\begin{equation} \label{equ:scale}
     s \geq - \frac{\ln(\frac{1}{P_{i}} - 1)}{2 - m_{i} -\frac{\sum_{j \neq i} m_j}{|\Omega| - 1}}.
\end{equation}
\subsection{Comparison with Different Loss Functions}
\begin{figure}
  \centering
  \includegraphics[width=1.0\linewidth]{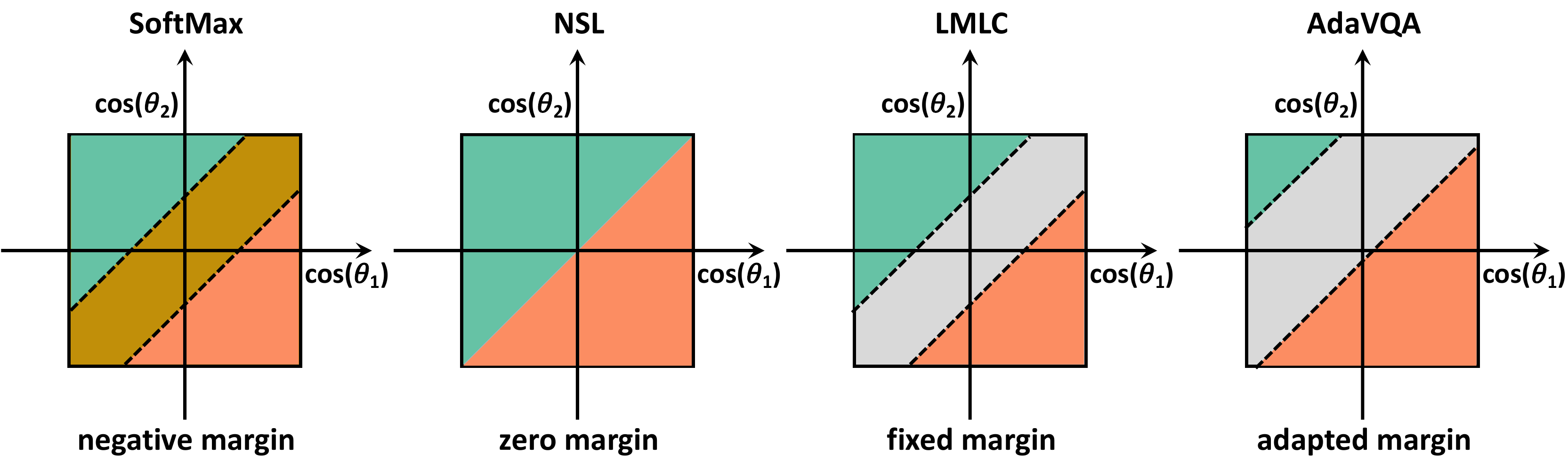}
  \caption{The visual comparison of decision boundary from different methods.}\label{fig:model}
\end{figure}

We consider the binary-class scenario for intuitively illustrating the decision boundary from different loss functions. As can be seen from Fig.~\ref{fig:model}, the decision boundary can be negative of the plain SoftMax one, which is enlarged to be equal to zero of the NSL loss~\cite{cosface}. The LMLC~\cite{cosface} defines a fixed margin for different classes, which is not suitable in our case for overcoming the language prior problem. Regarding our AdaVQA, the sparser answer (green one) engages a smaller feature space while the more frequent answer (orange one) engages a larger feature space.
\section{Experiments}\label{experiments}
\subsection{Experimental Setup}
\textbf{Datasets.}
To validate the effectiveness of the proposed loss function, we conducted extensive experiments on the two VQA-CP datastes: VQA-CP v2 and VQA-CP v1~\cite{vqacp}, which are two public benchmarks for estimating the models' capability of overcoming the language prior problem in VQA.

\noindent \textbf{Evaluation Metric.}
The standard metric in VQA is adopted for evaluation~\cite{vqa1}. That is, for each predicted answer $a$, the accuracy is computed as,
\begin{equation}\label{equ:metric}
  Acc_a = \text{min} (1, \frac{\#\text{humans that provide answer $a$}}{3}).
\end{equation}
Note that each question is answered by ten annotators, and this metric takes the disagreement in human answers into consideration~\cite{vqa1,vqa2}. In addition, the answers are divided into three categories: \emph{yes/no}, \emph{number} and \emph{other}.

\noindent  \textbf{Implementation Details.}
We empirically applied our AdaVQA to three baselines: Strong-BL~\cite{strong}, Counter~\cite{count} and UpDn~\cite{updown}. The prevailing visual attention (i.e., focusing on salient image regions with respect to given questions) is employed by these methods. And UpDn is a popular backbone for estimating the alleviation effect of language priors from the existing methods. Different from most prior methods overcoming the language prior problem in VQA, for all the three baselines, we simply replaced the original loss with our AdaVQA and did NOT change any other settings, such as embedding size, learning rate, optimizer, and batch size.

\subsection{Experimental Results}\label{results}
\begin{table}[htbp]
  \centering
  \caption{Accuracy comparisons with respect to different answer categories over the VQA-CP v2 dataset. Regarding the method category, the first group denotes plain approaches, the second group represents methods directly applied on the UpDn baseline, and the approaches from the last group are with our loss function. `$-$' and `$\dag$' denote the numbers are not available and our implementation, respectively. The best performance in current splits is highlighted in bold.}\label{tab:baseline-v2}
  \scalebox{1.0}{
  \begin{tabular}{lcccc}
    \toprule
    \multirow{2}{*}{Method}
                            & \multicolumn{4}{c}{VQA-CP v2 test}    \\
                              \cmidrule(lr){2-5}                     
                            & Y/N   & Num.  & Other & All           \\
    \midrule
    Question-Only [2015]    & 35.09 & 11.63 & 07.11 & 15.95         \\
    SAN [2016]              & 38.35 & 11.14 & 21.74 & 24.96         \\
    NMN [2016]              & 38.94 & 11.92 & 25.72 & 27.47         \\
    MCB [2016]              & 41.01 & 11.96 & 40.57 & 36.33         \\
    Strong-BL\dag [2017]    & 40.04 & 12.01 & 37.60 & 34.41         \\ 
    HAN [2018]              & 52.25 & 13.79 & 20.33 & 28.65         \\
    Counter\dag [2018]      & 41.01 & 12.98 & 42.69 & 37.67         \\
    UpDn [2018]             & 42.27 & 11.93 & 46.05 & 39.74         \\
    UpDn\dag [2018]         & 49.78 & 14.07 & 43.42 & 40.79         \\
    \midrule
    GVQA [2018]             & 57.99 & 13.68 & 22.14 & 31.30         \\
    AdvReg [2018]           & 65.49 & 15.48 & 35.48 & 41.17         \\
    Rubi [2019]             & 68.65 & 20.28 & 43.18 & 47.11         \\
    LMH [2019]              & -     & -     & -     & 52.05         \\
    LMH\dag [2019]          & 70.29 & 44.10 & 44.86 & 52.15         \\
    HINT [2019]             & 67.27 & 10.61 & 45.88 & 46.73         \\
    SCR [2019]              & 72.36 & 10.93 & \bf{48.02} & 49.45         \\
    VGQE [2020]             & 66.35 & 27.08 & 46.77 & 50.11         \\
    CSS [2020]              & 43.96 & 12.78 & 47.48 & 41.16    \\
    Decomp-LR [2020]        & 70.99 & 18.72 & 45.57 & 48.87         \\
    \midrule
    Strong-BL+Ours          & 62.70 & 21.64 & 35.33 & 41.21         \\
    Counter+Ours            & 61.00 & 53.22 & 43.17 & 49.90         \\
    UpDn+Ours               & \bf{72.47} & \bf{53.81} & 45.58 & \bf{54.67}    \\
    \bottomrule
  \end{tabular}
  }
\end{table}
\begin{table}[htbp]
  \centering
  \caption{Accuracy comparisons with respect to different answer categories over the VQA-CP v1 dataset. `$\dag$' denotes our implementation. The best performance in current splits is highlighted in bold.}\label{tab:baseline-v1}
  \scalebox{1.0}{
  \begin{tabular}{lcccc}
    \toprule
    \multirow{2}{*}{Method}
                            & \multicolumn{4}{c}{VQA-CP v1 test}    \\
                              \cmidrule(lr){2-5}
                            & Y/N   & Num.  & Other & All           \\
    \midrule
    Question-only [2015]    & 35.72 & 11.07 & 08.34 & 20.16         \\
    SAN [2016]              & 35.34 & 11.34 & 24.70 & 26.88         \\
    NMN [2016]              & 38.85 & 11.23 & 27.88 & 29.64         \\
    MCB [2016]              & 37.96 & 11.80 & 39.90 & 34.39         \\
    Strong-BL\dag [2017]    & 40.04 & 12.01 & 37.60 & 34.41         \\ 
    Counter\dag [2018]      & 41.01 & 12.98 & 42.69 & 37.67         \\
    UpDn\dag [2018]         & 43.76 & 12.49 & 42.57 & 38.02         \\
    GVQA [2018]             & 64.72 & 11.87 & 24.86 & 39.23         \\
    AdvReg [2018]           & 74.16 & 12.44 & 25.32 & 43.43         \\
    LMH\dag [2019]          & 76.61 & 29.05 & \bf{43.38} & 54.76    \\
    \midrule
    Strong-BL+Ours          & 72.64 & 27.61 & 36.11 & 49.84         \\
    Counter+Ours            & 72.01 & \bf{49.28} & 42.60 & 55.92    \\
    UpDn+Ours               & \bf{91.17} & 41.34 & 39.38 & \bf{61.20}\\
    \bottomrule
  \end{tabular}
  }
\end{table}
\subsubsection{Overall Accuracy} 
We conducted extensive experiments on two VQA-CP datasets, i.e., VQA-CP v2 and VQA-CP v1, and illustrated the results in Tab.~\ref{tab:baseline-v2} and Tab.~\ref{tab:baseline-v1}, respectively. The main observations from these two tables are as follows:
\begin{itemize}[leftmargin=*]
    \item Our method achieves the best on the two benchmark datasets except for the \emph{Other} answer category (SCR~\cite{counterfactual} and LMH~\cite{lmh} outperform ours on VQA-CP v2 and VQA-CP v1, respectively.). One possible reason for this is that most questions under this answer category are starting with `what ...', and the answers are therefore much more diverse than other categories. Accordingly, less language prior would be introduced under this category since the answers are less biased.
    \item For all the three baselines, i.e., Strong-BL, Counter and UpDn, when applying our loss function, a significant performance improvement (15\% on average) can be observed. For example, on the VQA-CP v2 dataset, Counter+Ours achieves an absolute performance gain of 12.23\% compared with Counter; and on the VQA-CP v1 dataset, UpDn+Ours outperforms the baseline UpDn with 23.18\%.
    \item Compared with other methods whose backbone model is also UpDn on the VQA-CP v2 dataset, our method (UpDn+Ours) surpasses them with a large margin, especially for the three newly accepted approaches VGQE, CSS and Decomp-LR.
\end{itemize}
\begin{figure}
  \centering
  \includegraphics[width=1.0\linewidth]{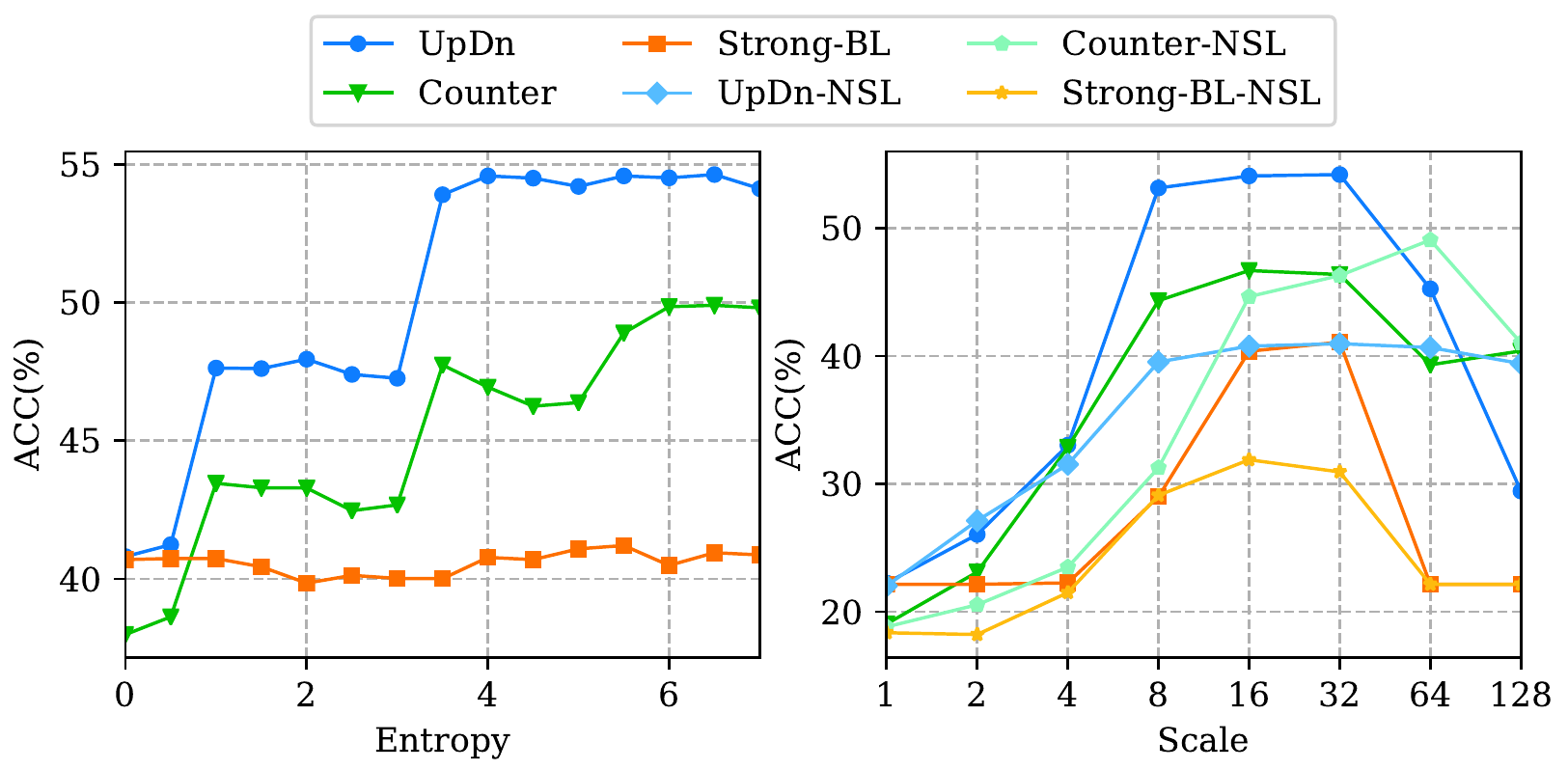}
  \caption{Test set accuracy with respect to different question type entropy and scales on the VQA-CP v2 dataset.}\label{fig:line-v2}
\end{figure}

\begin{table}
  \centering
  \caption{Effectiveness validation of the proposed loss function on three baselines over the VQA-CP v2 dataset.}\label{tab:ablation-v2}
  \scalebox{0.95}{
  \begin{tabular}{lcccc}
    \toprule
    Model               & Margin            & Strong-BL & Counter   & UpDn   \\
    \midrule
    Baseline            & -                 & 34.41     & 37.67     & 40.79  \\
    NSL                 & -                 & 31.89     & 49.08     & 40.97  \\
    \midrule
    \multirow{5}{*}{Fixed Margin}       
                        & 0.1               & 34.42     & 38.03     & 40.62  \\
                        & 0.3               & 34.38     & 38.21     & 40.68  \\
                        & 0.5               & 34.29     & 38.07     & 40.88  \\
                        & 0.7               & 34.22     & 38.15     & 40.98  \\
                        & 0.9               & 34.38     & 38.10     & 41.02  \\
    \midrule
    Adapted Margin      & \emph{adapted}    & 41.21     & 49.90     & 54.67 \\ 
    \bottomrule
  \end{tabular}}
\end{table}

\subsubsection{Ablation Study} \label{exp:ablation}
For a deeper understanding of our AdaVQA, we further provided the detailed ablation study on the VQA-CP v2 dataset. From the results in Tab.~\ref{tab:ablation-v2} and Fig.~\ref{fig:line-v2}, we can see that:
\begin{itemize}[leftmargin=*]
\item As shown in Tab.~\ref{tab:ablation-v2}, when using the $L2$ normalization on both the weight vector and the feature vector, the de facto NSL, the results are inconsistent amongst different methods. For example, the Counter with NSL surpasses the baseline with 11.41\%, while Strong-BL with NSL even leads to a little bit deterioration of the performance.
\item We then leveraged the fixed margin to yield effective feature discrimination, where the margins are tuned from 0.1 to 0.9 with a step size of 0.2. However, the results are not favorable (the improvement is limited), which validates the evidence that a fixed margin is not suitable for overcoming the language prior problem. In contrast, when replacing the fixed margin with our adapted one, the model can substantially outperform the one equipped with fixed margins, which additionally proves the superiority of AdaVQA.
\item We investigated the different influence of the question type entropy and scales in Fig.~\ref{fig:line-v2}. From this figure, we found that 1) an appropriate entropy should be considered, since a very large one will not bring more gains or even cause performance degradation; and 2) too small scales are not sufficient for learning the feature space, while too large ones will also lead to unsatisfactory results. 
\end{itemize}

\subsubsection{Case Study}
\begin{figure}
  \centering
  \includegraphics[width=1.0\linewidth]{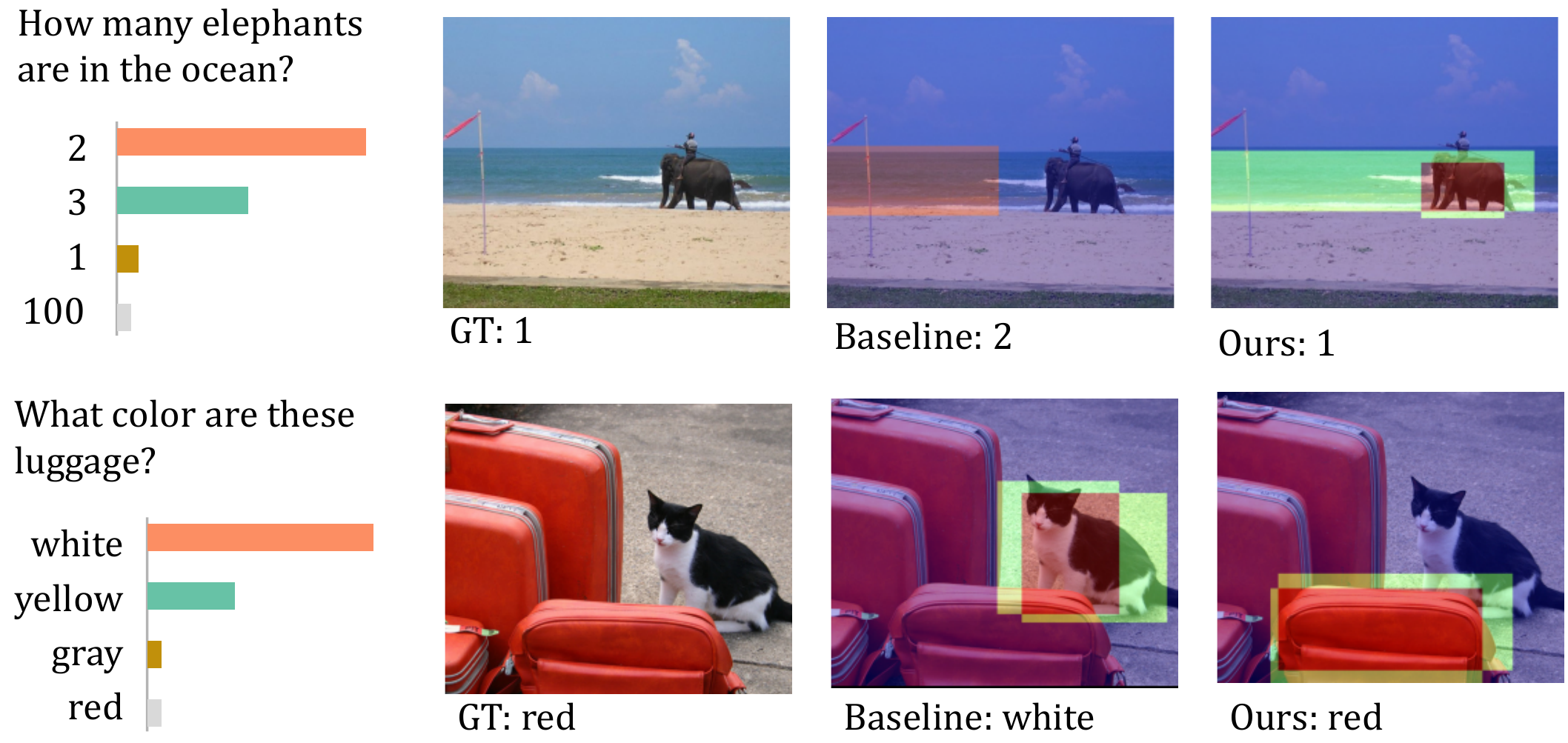}
  \caption{Visualization of the baselines method (UpDn) with and without our proposed loss function. The key answer distributions under the question's corresponding question type are illustrated on the leftmost column, followed by the attention maps from baseline and ours on the last two columns.}\label{fig:case-study}
\end{figure}

Finally, we visualized two successful cases from our method and illustrated them in Fig.~\ref{fig:case-study}. Regarding the first case, as the answer \emph{2} takes large proportion under the question type \emph{how many} in the train set, the baseline model thus yields an answer \emph{2} to this question. In contrast, our AdaVQA corrects this mistake, producing a more reasonable attention map (focusing on the elephant object). As for the second case, the baseline model wrongly predicting the answer \emph{white} mainly attribute to \emph{white} being more frequent under the question type \emph{what color} in the train set. Besides, too much attention has been paid to the cat region, which contributes another reason to the incorrect answer, as the fur of the cat is white and black. On the contrary, our AdaVQA can guide the model to pay more attention to the target object - luggage, leading to the correct answer \emph{red}.

\section{Conclusion}\label{conclusion}
We propose to tackle the language prior problem in VQA from the angle of the answer feature space learning, which has not been explored in previous studies. To achieve this goal, we design an  adapted margin cosine loss to discriminate the answers via properly characterizing the answer feature space. Concretely, for the given question, the frequent and sparse answers with respect to the corresponding question type are learned to span broader and tighter on the angular space, respectively. Extensive experiments over two benchmarks validate the effectiveness of the proposed loss function on three baselines.     

Nevertheless, this work does not present an advanced model for pursuing the SOTA results in VQA. We instead expect that, with this adapted margin cosine loss, future research can focus more on the enhancement of the visual reasoning capability without too much influence of the language prior problem. 

\newpage
% \input{segments/appendix}

% \begin{proof}
% \end{proof}

% \begin{algorithm}[tb]
% \caption{Example algorithm}
% \label{alg:algorithm}
% \textbf{Input}: Your algorithm's input\\
% \textbf{Parameter}: Optional list of parameters\\
% \textbf{Output}: Your algorithm's output
% \begin{algorithmic}[1] %[1] enables line numbers
% \STATE Let $t=0$.
% \WHILE{condition}
% \STATE Do some action.
% \IF {conditional}
% \STATE Perform task A.
% \ELSE
% \STATE Perform task B.
% \ENDIF
% \ENDWHILE
% \STATE \textbf{return} solution
% \end{algorithmic}
% \end{algorithm}

%\section*{Acknowledgments}
%\appendix
%\section{\LaTeX{} and Word Style Files}\label{stylefiles}

\bibliographystyle{reference/named}
\bibliography{reference/ijcai21}

\end{document}

\maketitle

% \begin{abstract}
%   The
% \end{abstract}

% \input{segments/introduction}
% \input{segments/related_work}
\section{Proposed AdaVQA}\label{model}
\subsection{Formulation of Partial Derivatives}
Let $p_i = s (\cos{\theta_{i}} - m_i) = s(\frac{\mathbf{W}_i^T}{||\mathbf{W}_i||_2} \cdot \frac{\mathbf{x}}{||\mathbf{x}||_2} - m_i)$, and $\hat{p}_i = \frac{\exp{p_i}}{\sum_{j=1}^{|\Omega|}\exp{p_j}}$. According to the derivative of the softmax function, we have, 
\begin{equation}\label{equ:derivative_softmax}
  \frac{\partial \hat{p}_i}{\partial p_i} =
  \begin{cases}
    \hat{p}_i (1- \hat{p}_j), & \mbox{if } i == j \\
    - \hat{p}_i \hat{p}_j. & \mbox{otherwise}.
  \end{cases}
\end{equation}
Consequently, 
\begin{equation}
\begin{aligned}
    \frac{\partial L_{AdaVQA}}{\partial p_i} &= -\sum_{i=1}^{|\Omega|} a_i \frac{\partial \log \hat{p}_i}{\partial p_i} \\
    &= -\sum_{i=1}^{|\Omega|} a_i \frac{1}{\hat{p}_i} \frac{\partial \hat{p}_i}{\partial p_i} \\
    &= a_i (\hat{p}_i - 1)   + \sum_{j \neq i}^{|\Omega|} a_j \hat{p}_i   \\
    &= \sum_{j=1}^{|\Omega|} a_j \hat{p}_i - a_i.  \\
    \end{aligned}
\end{equation}
In addition, the derivative of vector norms is given by,
\begin{equation}
    \frac{\partial \frac{\mathbf{x}}{||\mathbf{x}||_2}}{\partial \mathbf{x}} = \frac{||\mathbf{x}||_2 \mathbf{I} - \mathbf{x}\mathbf{x}^T}{||\mathbf{x}||_2^3},
\end{equation}
which gives, 
\begin{equation}
\begin{aligned}
    & \frac{\partial L_{AdaVQA}}{\partial \mathbf{x}} = \sum_{i=1}^{|\Omega|} \frac{\partial L_{AdaVQA}}{\partial p_i} \cdot \frac{\partial p_i}{\partial \mathbf{x}} \\
    &= \sum_{i=1}^{|\Omega|} (\sum_{j=1}^{|\Omega|} a_j \hat{p}_i - a_i) \cdot s \cdot  \frac{||\mathbf{x}||_2 \mathbf{I} - \mathbf{x}\mathbf{x}^T}{||\mathbf{x}||_2^3} \frac{\mathbf{W}_i}{||\mathbf{W}_i||_2}.
    \end{aligned}
\end{equation}
Similarly, the partial derivative for weight vector $\mathbf{W}_i$ can be obtained by,
\begin{equation}
\begin{aligned}
    &\frac{\partial L_{AdaVQA}}{\partial \mathbf{W}_i} = \sum_{i=1}^{|\Omega|} \frac{\partial L_{AdaVQA}}{\partial p_i} \cdot \frac{\partial p_i}{\partial \mathbf{W}_i} \\
    &= \sum_{i=1}^{|\Omega|} (\sum_{j=1}^{|\Omega|} a_j \hat{p}_i - a_i) \cdot s \cdot  \frac{||\mathbf{W}_i||_2 \mathbf{I} - \mathbf{W}_i \mathbf{W}_i^T}{||\mathbf{W}_i||_2^3} \frac{\mathbf{x}}{||\mathbf{x}||_2}.
\end{aligned}
\end{equation}
Based on this, the partial derivatives of other parameters can also be deduced through the chain rule.

\subsection{Proof for the Scaling Factor s}
Without loss of generality, let $P_{i}$ denotes the expected minimum of the answer class $i$. In the ideal formulation, the $\theta_i$ between answer  $i$ and weight vector $\mathbf{W}_i$ should be $0$, while that for others answers should be $180$. We then have:
\begin{equation*}
\begin{aligned}
    \frac{\exp{s (1 - m_{i})}}{\sum_{j \neq i} \exp{s (-1 - m_{j})} + \exp{s (1 - m_{i})}} &\geq P_{i}, \\
    1 + \frac{\sum_{j \neq i} \exp{s (-1 - m_{j})}}{\exp{s (1 - m_{i})}} &\leq \frac{1}{P_{i}}, \\
    1 + \frac{\sum_{j \neq i} \exp{s (m_{j})}}{\exp{s (2 - m_{i})}} &\leq \frac{1}{P_{i}}. \\
\end{aligned}
\end{equation*}
On the basis of Jensen's inequality,
\begin{equation*}
\begin{aligned}
    \sum_{j \neq i} \exp{s (m_{j})} &= \frac{|\Omega| - 1}{|\Omega| - 1} \sum_{j \neq i} \exp{s (m_{j})}, \\
    &\geq (|\Omega| - 1) \exp{s(\frac{\sum_{j \neq i} m_j}{|\Omega| - 1})}.
\end{aligned}
\end{equation*}
Accordingly, we obtain,
\begin{equation*}
\begin{aligned}
    1 + \frac{(|\Omega| - 1) \exp{s(\frac{\sum_{j \neq i} m_j}{|\Omega| - 1})}}{\exp{s (2 - m_{i})}} &\leq \frac{1}{P_{i}}, \\
    \exp{s(\frac{\sum_{j \neq i} m_j}{|\Omega| - 1} + m_{i} -2)} &\leq \frac{1}{P_{i}} - 1.\\
\end{aligned}
\end{equation*}
Finally, we have,
\begin{equation} \label{equ:scale}
    s \geq - \frac{\ln(\frac{1}{P_{i}} - 1)}{2 - m_{i} -\frac{\sum_{j \neq i} m_j}{|\Omega| - 1}}.
\end{equation}

\section{Experimental Settings}
\subsection{Datasets}
We validated the effectiveness of our proposed loss function on the two VQA-CP datastes: VQA-CP v2 and VQA-CP v1~\cite{vqacp}, which are two public benchmarks for estimating the models' capability of overcoming the language prior problem in VQA and are the unbiased splits of the traditional VQA datasets\cite{vqa1,vqa2}. The VQA-CP v2 and VQA-CP v1 datasets consist of $\sim$122K images, $\sim$658K questions and $\sim$6.6M answers, and $\sim$122K images, $\sim$370K questions and $\sim$3.7M answers, respectively. And each question is answered by ten annotators. Moreover, the answer distribution of each question type is significantly different between the training and the testing sets in the VQA-CP datasets. For example, in VQA-CP v1, \emph{1} and \emph{3} take about 50\% of the answers under the question type \emph{how many} in the training set, while \emph{2} and \emph{4} hold around 75\% in the testing set. 

\subsection{Baselines}
We mainly tested our method over the following three baselines: Strong-BL~\cite{strong}, Counter~\cite{counterfactual} and UpDn~\cite{updown}. Thereinto, Strong-BL utilizes the pre-trained grid-based image features directly produced from CNNs~\cite{resnet}, while Counter and UpDn use the pre-trained region-based image features from faster-RCNN~\cite{rcnn}.

\textbf{Strong-BL}~\cite{strong} first leverages two stacked convolutional layers to obtain the attention weights for each equal-sized image grid. Afterwards, it fuses the attentive image features with the question features from RNNs~\cite{gru} in a vector additive fashion. 

\textbf{Counter}~\cite{count} is an upgraded model of the Strong-BL, introducing a counting module to enable robust counting from object proposals. 

\textbf{UpDn}~\cite{updown} firstly leverages the pre-trained object detection networks to obtain salient object features for VQA. And then it employs a simple attention network to focus on the most important objects which are highly related with the given question.
\section{Experimental Results}\label{results}
\subsubsection{Ablation Study.} 
\begin{figure}
  \centering
  \includegraphics[width=1.0\linewidth]{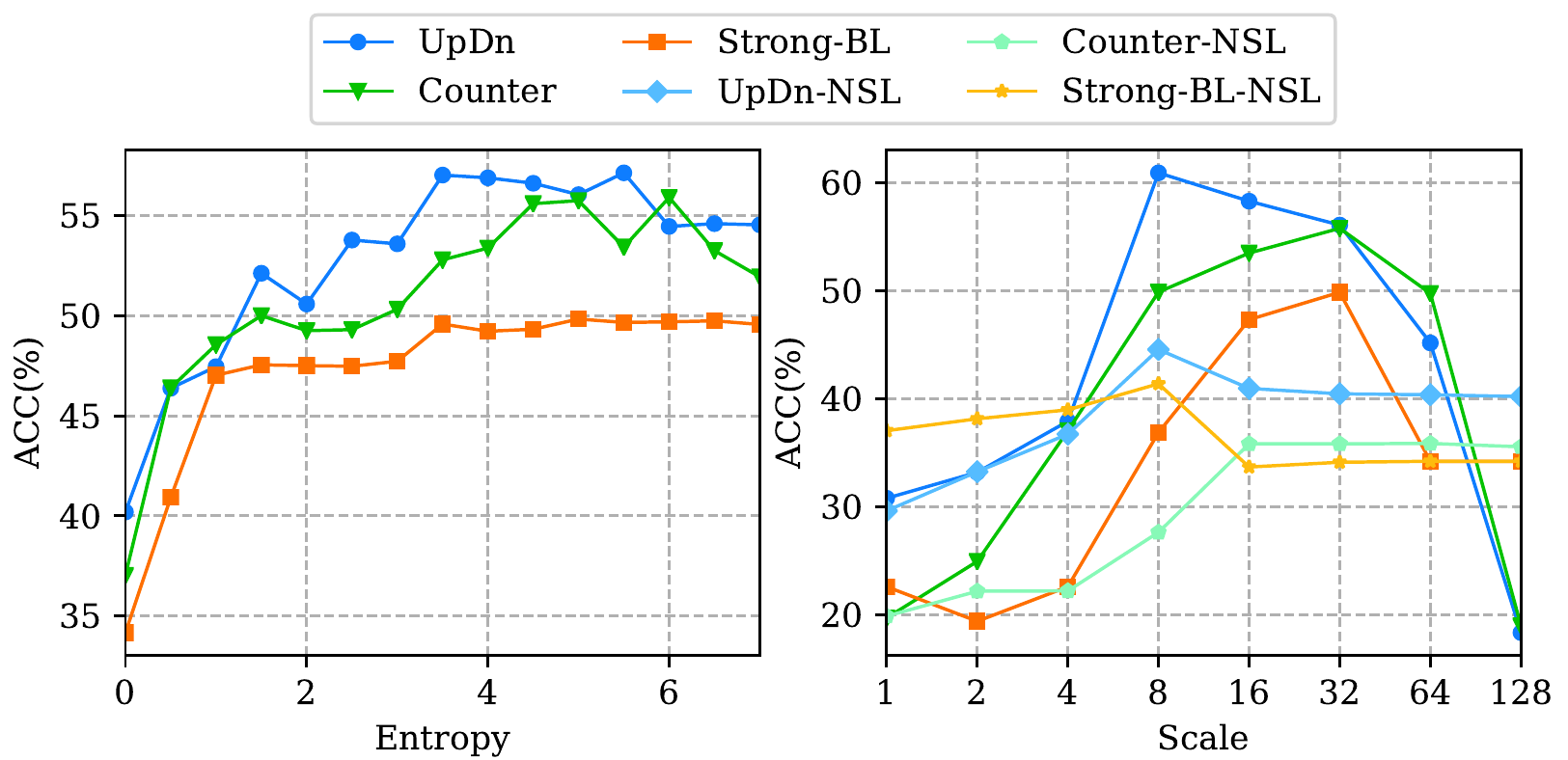}
  \caption{Test set accuracy with respect to different question type entropy and scales on the VQA-CP v1 dataset.}\label{fig:line-v1}
\end{figure}

\begin{figure*}
  \centering
  \includegraphics[width=1.0\linewidth]{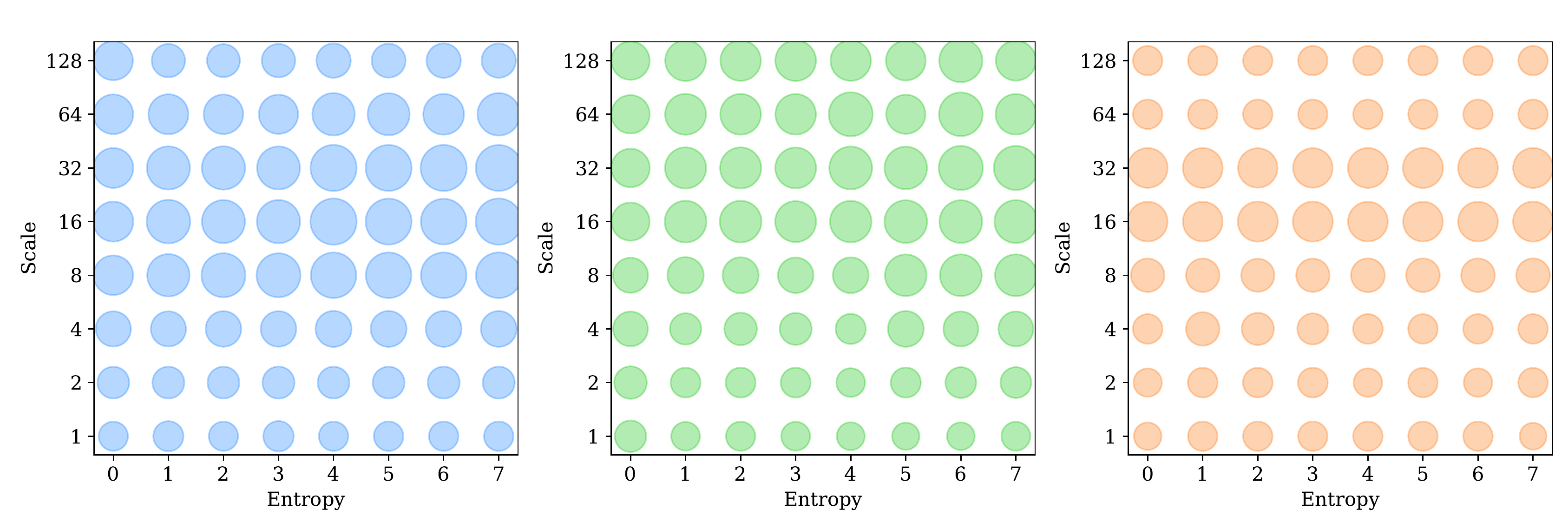}
  \caption{Test set accuracy with respect to different question type entropy as well as different scales on the VQA-CP v2 dataset.}\label{fig:bubble}
\end{figure*}

The detailed analysis of the margin on the VQA-CP v1 is provided in Fig.~\ref{fig:line-v1} and Tab.~\ref{tab:ablation-v1}. We also demonstrate the elaborate ACC results with respect to both question type entropy and number of scales in Fig.~\ref{fig:bubble}.

\subsubsection{Answer Manifold Embedding.} 
\begin{figure}
  \centering
  \begin{minipage}{1.0\linewidth}
  \centering
  \includegraphics[width=1.0\linewidth]{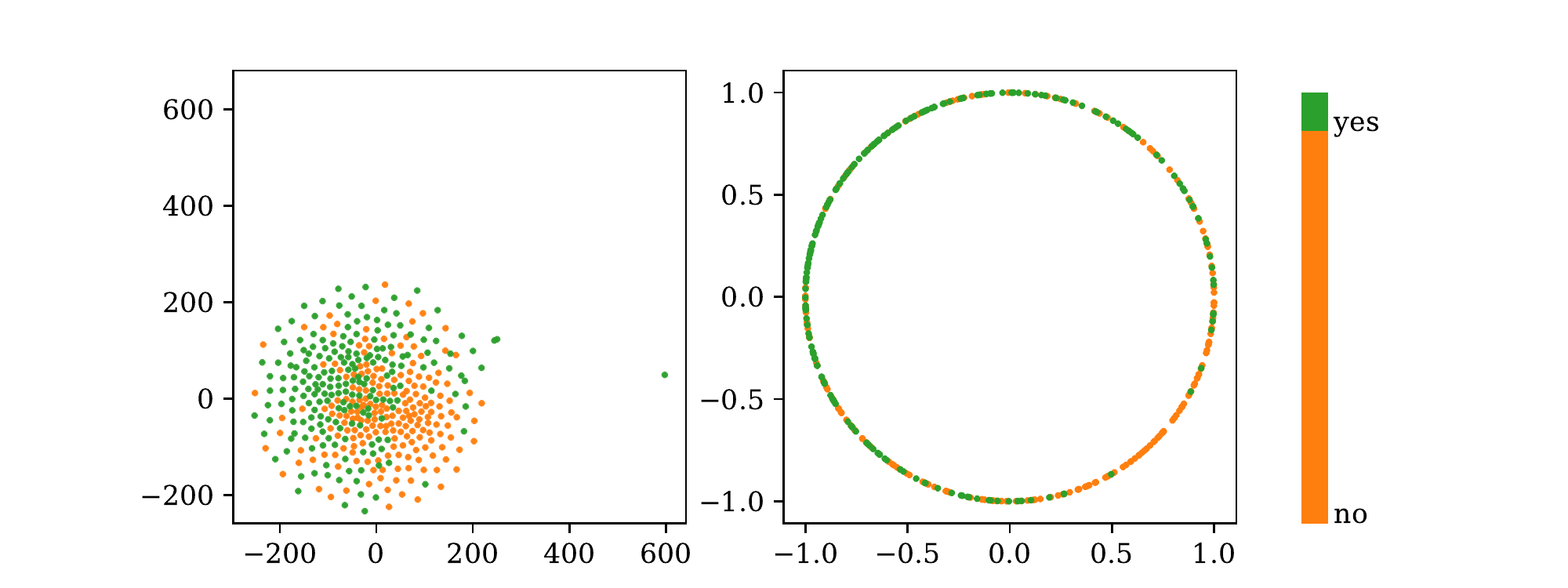}
  \subcaption{Answer manifold embedding of \emph{is this} question type in Euclidean and angular spaces from the baseline. }
  \end{minipage}
  
  \begin{minipage}{1.0\linewidth}
  \centering
  \includegraphics[width=1.0\linewidth]{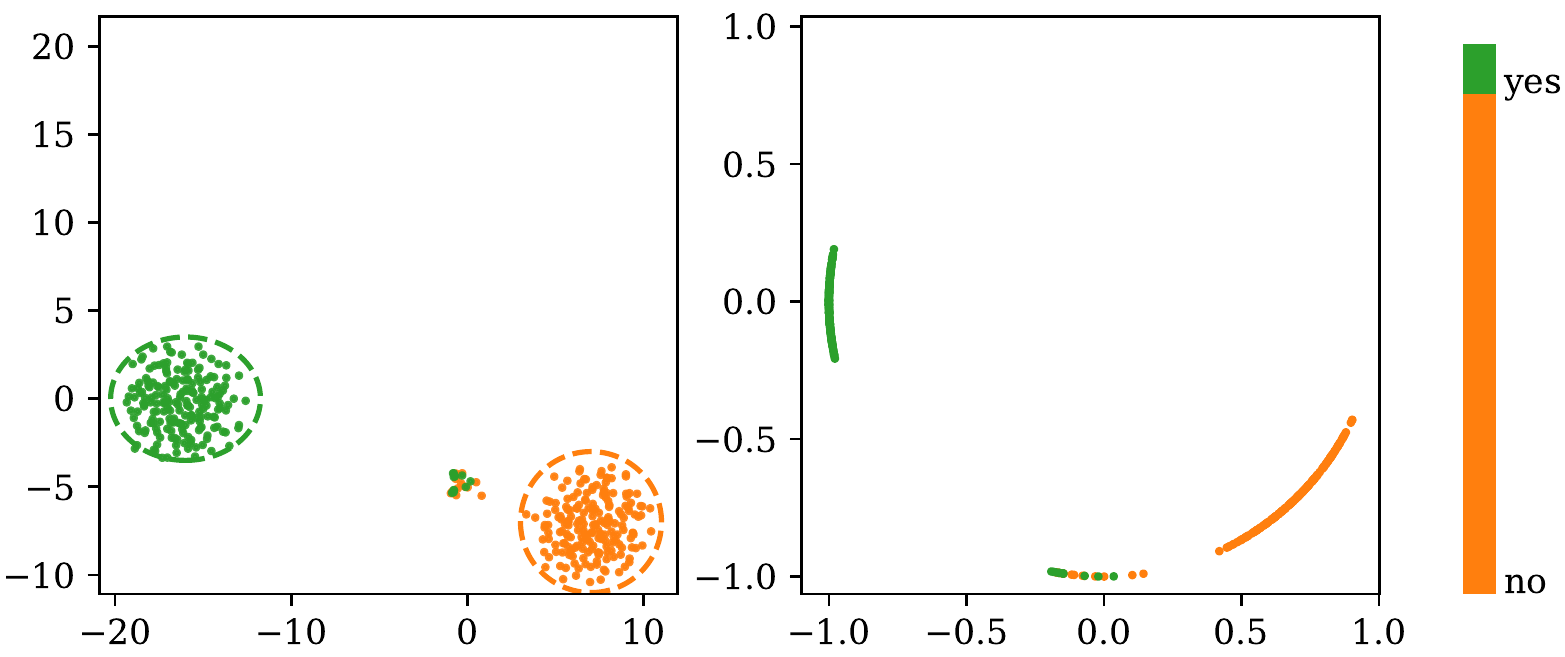}
  \subcaption{Answer manifold embedding of \emph{is this} question type in Euclidean and angular spaces from our method. }
  \end{minipage}
  \caption{Answer manifold embedding of \emph{is this} question type.}\label{fig:scatter-2}
\end{figure}

\begin{figure}
  \centering
  \begin{minipage}{1.0\linewidth}
  \centering
  \includegraphics[width=1.0\linewidth]{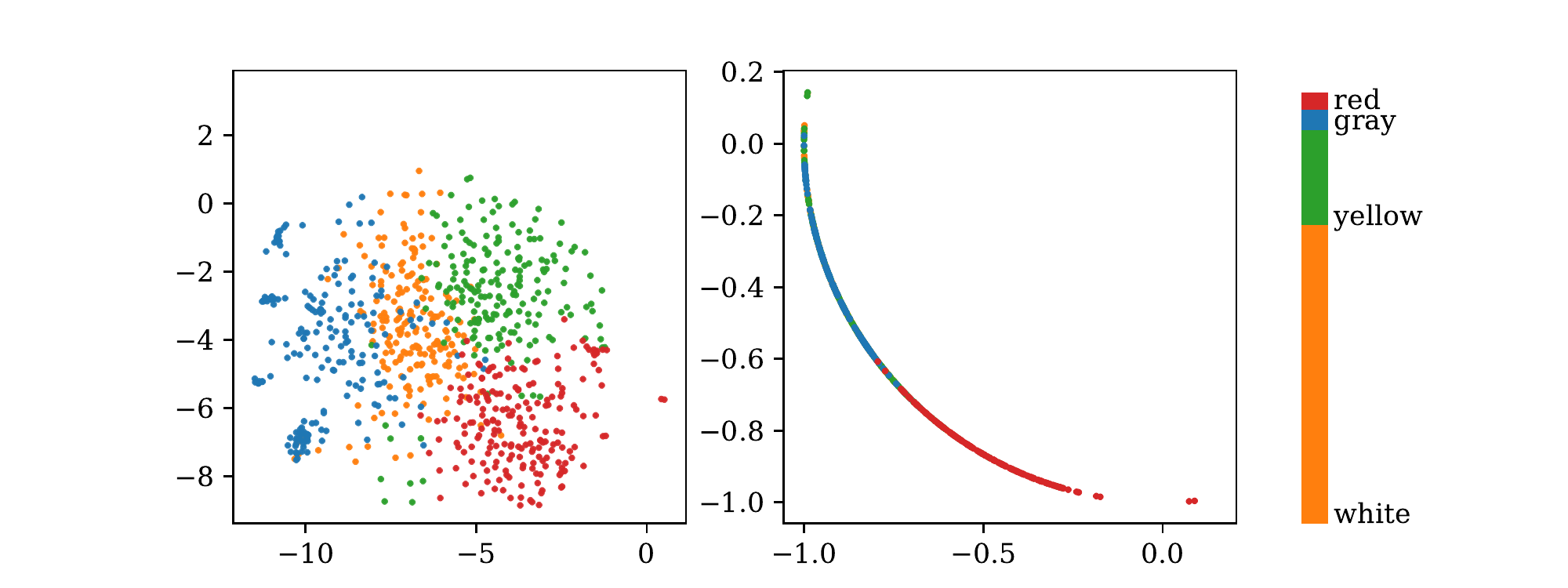}
  \subcaption{Answer embedding of \emph{what color is the} question type in Euclidean and angular spaces from the baseline. }
  \end{minipage}
  
  \begin{minipage}{1.0\linewidth}
  \centering
  \includegraphics[width=1.0\linewidth]{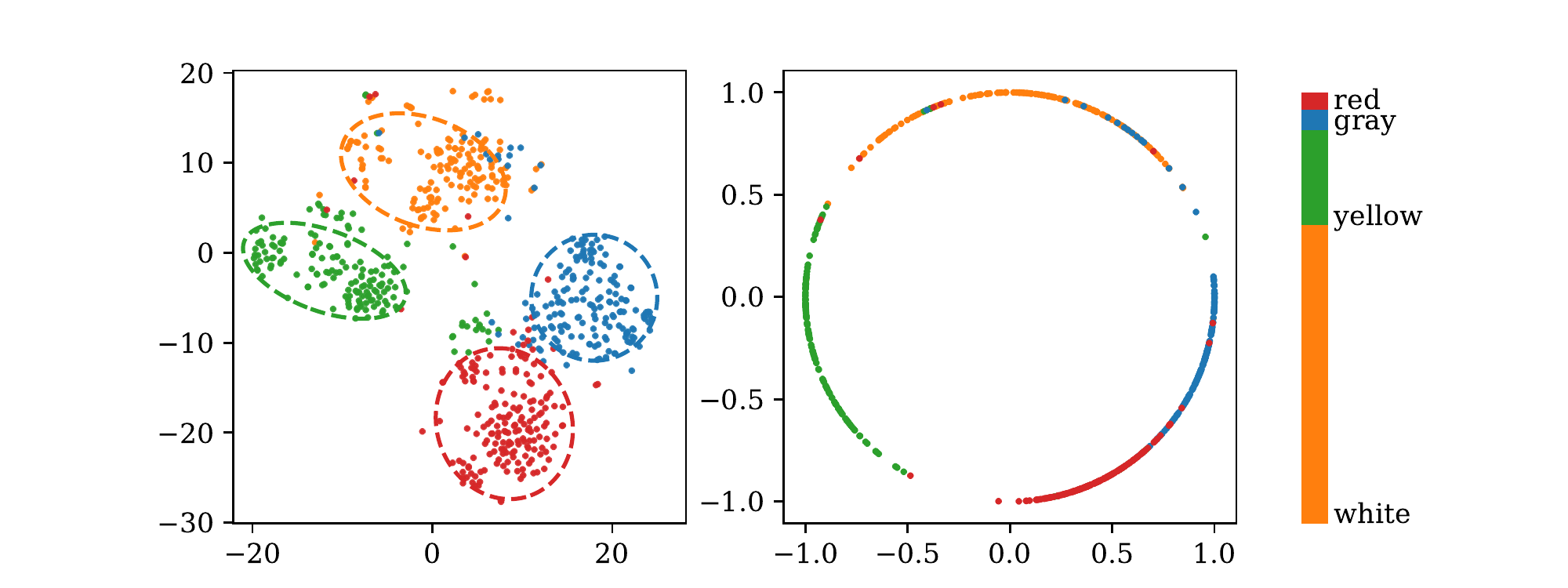}
  \subcaption{Answer manifold embedding of \emph{what color is the} question type in Euclidean and angular spaces from our method. }
  \end{minipage}
  \caption{Answer embedding of \emph{what color is the} question type.}\label{fig:scatter-4}
\end{figure}

The answer embedding results from two more question types, i.e., \emph{is this} and \emph{what color is the} are shown in Fig.~\ref{fig:scatter-2} and Fig.~\ref{fig:scatter-4}, respectively. 

\subsubsection{Case Study.} 
\begin{figure*}
  \centering
  \includegraphics[width=1.0\linewidth]{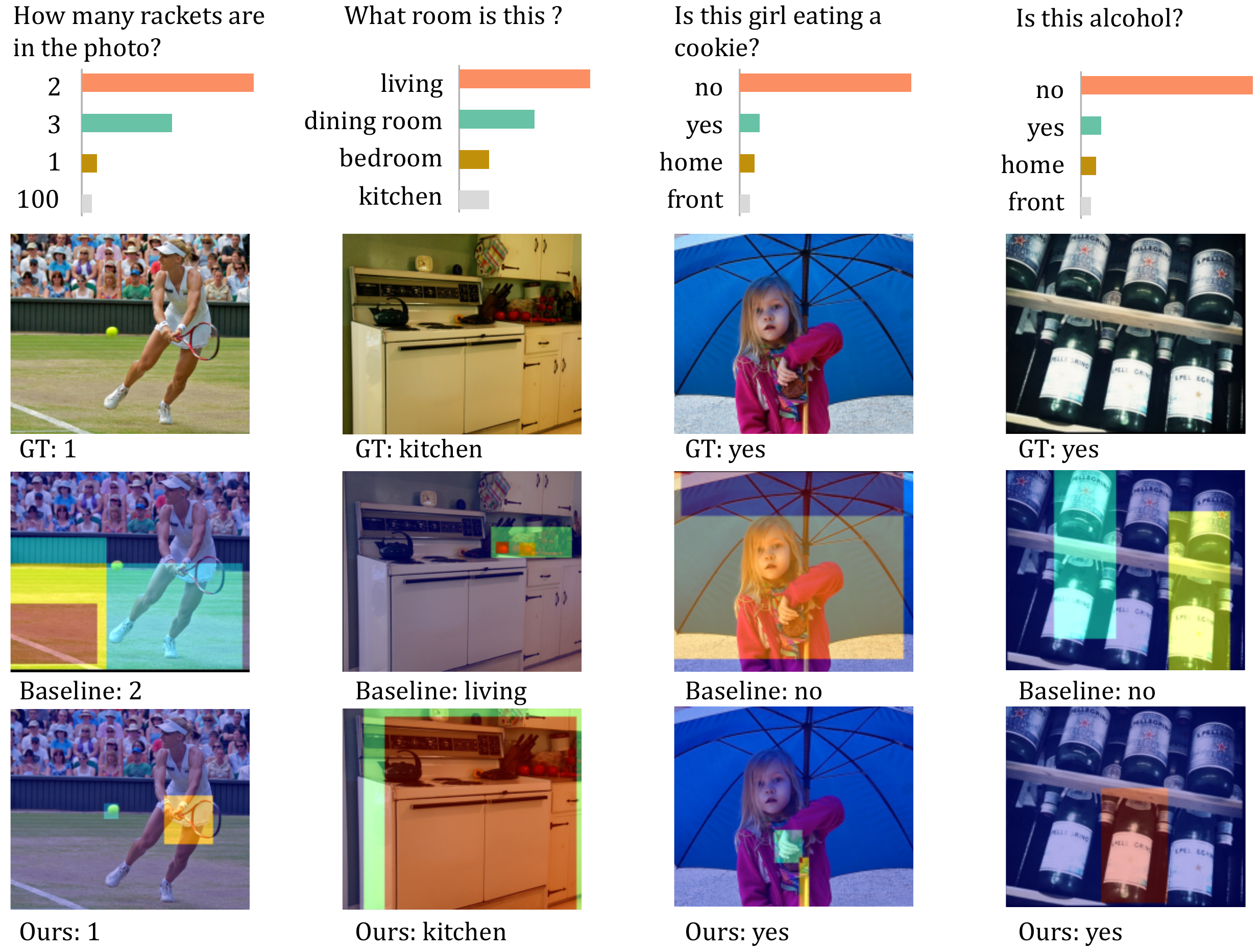}
  \caption{Visualization of the baselines method (UpDn) with and without our proposed loss function. The key answer distributions under the question's corresponding question type are illustrated on the second row, followed by the attention maps from baseline and ours on the last two rows.}\label{fig:case-study}
\end{figure*}

More attention map cases can be found at Fig.~\ref{fig:case-study}.
\begin{table}
  \centering
  \caption{Effectiveness validation of the proposed loss function on three baselines over the VQA-CP v1 dataset.}\label{tab:ablation-v1}
  \scalebox{0.95}{
  \begin{tabular}{lcccc}
    \toprule
    Model               & Margin            & Strong-BL & Counter   & UpDn   \\
    \midrule
    Baseline            & -                 & 34.19     & 37.08     & 41.34  \\
    NSL                 & -                 & 41.38     & 35.86     & 44.55  \\
    \midrule
    \multirow{5}{*}{Fixed Margin}       
                        & 0.1               & 33.79     & 37.54     & 40.92  \\
                        & 0.3               & 33.90     & 37.34     & 40.31  \\
                        & 0.5               & 33.83     & 37.43     & 40.16  \\
                        & 0.7               & 34.38     & 37.72     & 40.84  \\
                        & 0.9               & 34.15     & 37.78     & 41.13  \\
    \midrule
    Adapted Margin      & \emph{adapted}    & 49.84     & 55.92     & 61.20 \\ 
    \bottomrule
  \end{tabular}}
\end{table}

% \begin{proof}
% \end{proof}

% \begin{algorithm}[tb]
% \caption{Example algorithm}
% \label{alg:algorithm}
% \textbf{Input}: Your algorithm's input\\
% \textbf{Parameter}: Optional list of parameters\\
% \textbf{Output}: Your algorithm's output
% \begin{algorithmic}[1] %[1] enables line numbers
% \STATE Let $t=0$.
% \WHILE{condition}
% \STATE Do some action.
% \IF {conditional}
% \STATE Perform task A.
% \ELSE
% \STATE Perform task B.
% \ENDIF
% \ENDWHILE
% \STATE \textbf{return} solution
% \end{algorithmic}
% \end{algorithm}

%\section*{Acknowledgments}
%\appendix
%\section{\LaTeX{} and Word Style Files}\label{stylefiles}

\bibliographystyle{reference/named}
\bibliography{reference/ijcai21}